\documentclass[runningheads]{llncs}
\usepackage{graphicx}

\usepackage{times}
\usepackage{xcolor}
\usepackage[utf8]{inputenc}
\usepackage[small]{caption}
\usepackage{graphicx}
\usepackage{subfigure}

\usepackage{amsmath}
\usepackage{amssymb}
\usepackage{booktabs}
\usepackage{multirow}
\begin{document}
\title{Cross-dataset Person Re-Identification Using Similarity Preserved Generative Adversarial Networks\thanks{The work described in this paper was supported by the grants from NSFC (No. U1611461), Science and Technology Program of Guangdong Province, China (No. 2016A010101012), and CAS Key Lab of Network Data Science and Technology, Institute of Computing
Technology, Chinese Academy of Sciences, 100190, Beijing, China.(No.CASNDST201703).}}
\titlerunning{Cross-dataset Person Re-Identification}
\author{Jianming Lv 
\and Xintong Wang}
\authorrunning{J. Lv et al.}
\institute{South China University of Technology, Guangzhou, China  \\
\email{\textbf{jmlv@scut.edu.cn}}\textbf{,} \email{\textbf{w.xintong@mail.scut.edu.cn}}}
\maketitle              % typeset the header of the contribution
\begin{abstract}
Person re-identification (Re-ID) aims to match the image frames which contain the same person in the surveillance videos. Most of the Re-ID algorithms conduct supervised training in some small labeled datasets, so directly deploying these trained models to the real-world large camera networks may lead to a poor performance due to underfitting. The significant difference between the source training dataset and the target testing dataset makes it challenging to incrementally optimize the model. To address this challenge, we propose a novel solution by transforming the unlabeled images in the target domain to fit the original classifier by using our proposed similarity preserved generative adversarial networks model, SimPGAN. Specifically, SimPGAN adopts the  generative adversarial networks with the cycle consistency constraint to transform the unlabeled images in the target domain to the style of the source domain. Meanwhile, SimPGAN uses the similarity consistency loss, which is measured by a siamese deep convolutional neural network, to preserve the similarity of the transformed images of the same person. Comprehensive experiments based on multiple real surveillance datasets are conducted, and the results show that our algorithm is  better than the state-of-the-art cross-dataset unsupervised person Re-ID algorithms.
\keywords{Cross-dataset  \and Person re-identification \and Similarity preserved.}
\end{abstract}
\section{Introduction}

\begin{figure}
\centering
{
\begin{minipage}[b]{0.56\textwidth}
\includegraphics[width=1\textwidth]{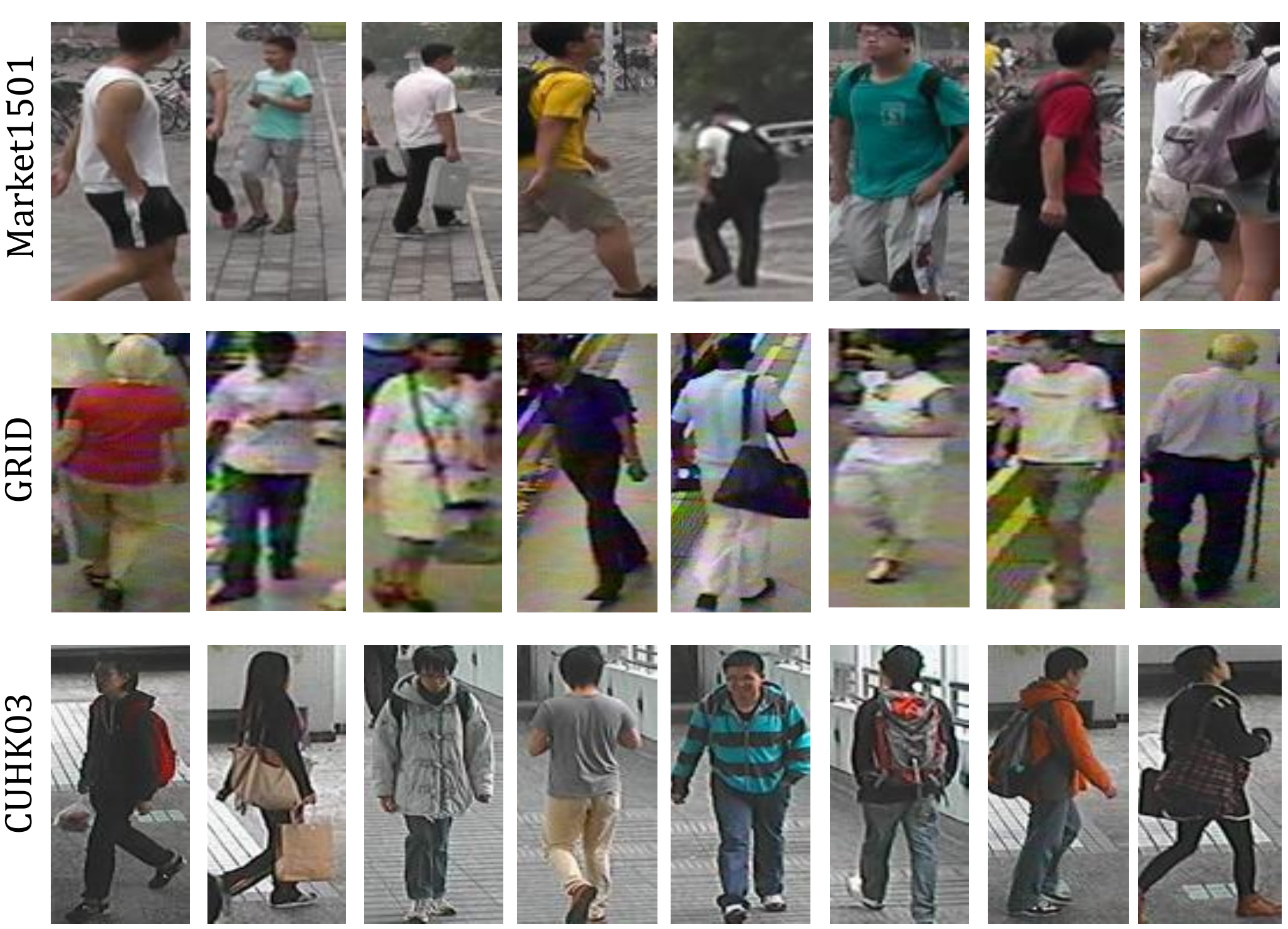}
\end{minipage}
}
\caption{Samples in different datasets.}
\label{fig:data_variation.}
\end{figure}

As one of the most important and challenging problems in the field of surveillance video analysis, person re-identification (Re-ID) aims to match the image frames which contain the same person in the surveillance videos. How to extract the view invariant features from the images and design a robust visual classifier to identify the persons is the core challenge of the Re-ID algorithms.

Due to the privacy problem regarding the collection of surveillance videos and the expensive cost of data labeling, most of the proposed Re-ID algorithms \cite{ref_proc25} \cite{ref_proc19}  conduct supervised learning on small labeled datasets. Directly deploying these trained models to the real-world large-scale camera networks may lead to a poor performance, because the images captured from different camera networks usually have totally different backgrounds, noise distributions, brightness and resolution  as shown in Fig.1. How to incrementally optimize the Re-ID algorithms based on the abundant unlabeled data collected from the target domain is a practical and extremely challenging problem.

To address this problem, some unsupervised algorithms  \cite{ref_proc10} \cite{ref_proc14} are proposed to extract view invariant features and measure the similarity of pedestrians based on the unlabeled dataset. Without the powerful supervision based on labels, this kind of pure unsupervised learning based algorithms working on one single dataset have a poor performance in most cases. Recently, a cross-dataset unsupervised transfer learning algorithm, named UMDL \cite{ref_proc18} ,  is proposed to make use of data samples from both labeled source datasets and the unlabeled target dataset to learn the view-invariant feature representation and similarity measurement by the dictionary learning mechanism.  UMDL gains much better performance than purely unsupervised algorithms, but is still much weaker than the state-of-the-art supervised algorithms based on the labeled dataset. Most of above algorithms try to incrementally optimize the visual classifier, which is pre-trained in the source dataset, to fit the new data in the target domain. However, without the labels in the target domain, it is hard to fine-tune the classifier to suit the source and target datasets simultaneously, which have diverse feature distributions.

We address this cross-dataset Re-ID challenge in a totally new direction in this paper. Instead of optimizing the classifier to fit the new data, we transform the unlabeled images in the target domain to fit the classifier by using our proposed similarity preserved generative adversarial networks model, SimPGAN. As shown in Fig.5 regarding an example from the real datsets, after the transformation, the features of  the images in the target datasets are projected into the features close to the distribution in the source dataset, and are easier to be processed by the visual classifier trained in the source dataset.

 The main contributions of this paper are summarized as follows:
\begin{itemize}
\item We propose a novel efficient solution, named SimPGAN,  to improve the performance of cross-dataset person Re-ID by transforming the unlabeled images in the target domain into the style of the source domain to fit the visual classifier transferred from the source domain.

\item SimPGAN adopts the  generative adversarial networks with cycle consistency constraint to avoid sharp change of the images after transformation. Meanwhile, SimPGAN uses the similarity consistency loss, which is measured by a siamese deep convolutional neural network, to preserve the similarity of the transformed images of the same person.

\item We conduct comprehensive experiments based on real datasets (Market1501\cite{ref_proc27}, CUHK01\cite{ref_proc23} , GRID\cite{ref_proc2}), which show that SimPGAN is  better than the state-of-the-art cross-dataset unsupervised transfer learning algorithm\cite{ref_proc18} .

\end{itemize}

 The rest of this paper is organized as follows. Section \ref{sec:related-work} reviews the related work of Re-ID.  Section \ref{sec:preliminaries} offers clear definitions of the problem about Re-ID in unlabeled dataset.  Section \ref{sec:model} presents  our proposed methods.  Section \ref{sec:experiment} evaluates the performance of this system by conducting experiments on real datasets. We conclude the work in Section \ref{sec:conclusions}.

\section{Related work}\label{sec:related-work}

\textbf{Supervised Learning}: Most existing person Re-ID models are supervised, and based on either invariant feature learning \cite{ref_proc11} ,  metric learning   \cite{ref_proc19}  or deep learning\cite{ref_proc25}  . However, in the practical deployment of Re-ID algorithms in large-scale camera networks, it is usually costly and unpractical to label the massive online surveillance videos to support supervised learning as mentioned in \cite{ref_proc18} .

\textbf{Unsupervised Learning}: In order to improve the effectiveness of the Re-ID algorithms towards large-scale unlabeled datasets, some unsupervised Re-ID methods \cite{ref_proc10} are proposed to learn cross-view identity-specific information from unlabeled datasets. However, due to the lack of the knowledge about identity labels, these unsupervised approaches usually yield much weaker performance compared to supervised learning approaches.

\textbf{Transfer Learning}: Recently, some cross-dataset transfer learning algorithms\cite{ref_proc13} \cite{ref_proc14} \cite{ref_proc18} \cite{ref_proc9}  are proposed to leverage the Re-ID models pre-trained in other labeled datasets to improve the performance  on target dataset. This type of Re-ID algorithms can be classified further into two categories: supervised transfer learning and unsupervised transfer learning  according to whether the label information of target dataset is given or not. Specifically, in the \textbf{supervised transfer learning} algorithms\cite{ref_proc9} \cite{ref_proc13} \cite{ref_proc14} , both of the source and the target datasets are labeled or have weak labels. \cite{ref_proc9}   is based on a SVM multi-kernel learning transfer strategy, and \cite{ref_proc14}  is based on cross-domain ranking SVMs. \cite{ref_proc13}   adopts multi-task metric learning models. On the other hand, the recently proposed cross-dataset \textbf{unsupervised transfer learning} algorithm for Re-ID, UMDL\cite{ref_proc18} , is totally different from above algorithms, and closer to real-world deployment environment where the target dataset is totally unlabeled. UMDL\cite{ref_proc18} transfers the view-invariant representation of a person's appearance from the source labeled dataset to the unlabeled target dataset by dictionary learning mechanisms, and gains much better performance. Although this kind of cross-dataset transfering algorithms are proved to outperform the purely unsupervised algorithms, they still have a long way to catch up the performance of the supervised algorithms, e.g. in the CUHK01\cite{ref_proc23}  dataset, UMDL \cite{ref_proc18}  can achieve $27.1\%$ rank-1 accuracy, while the accuracy of the  state-of-the-art supervised algorithms \cite{ref_proc20}  can reach $67\% $.

\textbf{Generative Adversarial Networks}: Generative adversarial networks (GAN) \cite{ref_proc4}  provide a way to learn deep representations without extensively annotated training data. Within the adversarial network, the generative model takes random inputs and tries to generate data samples, while the discriminative model learns to determine whether a sample is drawn from the model distribution or the data distribution. GAN is broadly used for image synthesis, image classification, image-to-image translation and super-resolution and achieves impressive results. Specifically, the image-to-image translation aims to learn the mapping between an input image and an output image using a set of aligned image pairs. More recently, the Pix2Pix framework
\cite{ref_proc8} uses a conditional generative adversarial network to learn a mapping from input to output images by using paired training data. Unlike these prior works, Zhu \cite{ref_proc29} learns the mapping without paired training examples using the cycle-consistent adversarial networks.

\section{Preliminaries}\label{sec:preliminaries}
\subsection{Problem Definition of Re-ID}
Given a surveillance image which contains a target pedestrian, the goal of a person Re-ID algorithm is to retrieve the surveillance videos for the image frames which contain the same person. For clarity of the problem definition, some notations describing Re-ID are introduced in this section.

Each surveillance image containing a pedestrian is denoted as $I_j$, which is cropped from an image frame of a surveillance video. The ID of the pedestrian in $I_j$ is denoted as $\Upsilon(I_j)$. Given any surveillance image $I_j$, the person Re-ID problem is to retrieve the images $\{I_k | \Upsilon(I_j) = \Upsilon(I_k)\}$, which contain the same person $\Upsilon(I_j)$.

The traditional strategy of person Re-ID is to train a classifier $\mathcal{C}$  based on visual features to judge whether two given images contain the same person. The output of the classifier is usually the similarity score, which measures the likehood that the two images contain the same person. The similarity score can be used to rank the image frames to retrieve the Re-ID results.

\subsection{Cross-Dataset Person Re-ID}

Like most of the traditional person Re-ID algorithms \cite{ref_proc11}, we can conduct supervised learning on some public labeled dataset (denoted as $\Omega_s$ below), which is usually of small size, to train a classifier $\mathcal{C}$. While directly deploying the trained $\mathcal{C}$ to a real-world unlabeled target dataset $\Omega_t$ collected from a large-scale camera network, it tends to have a poor performance, due to the significant difference between $\Omega_s$ and $\Omega_t$ as shown in the Fig.1.

How to effectively transfer the classifier trained in a small labeled source dataset to another unlabeled target datset is the fundamental challenging problem addressed in this paper.

\begin{figure}[!htbp]
\centering
\subfigure[]{
\begin{minipage}[b]{0.3\textwidth}
\includegraphics[width=1.1\textwidth]{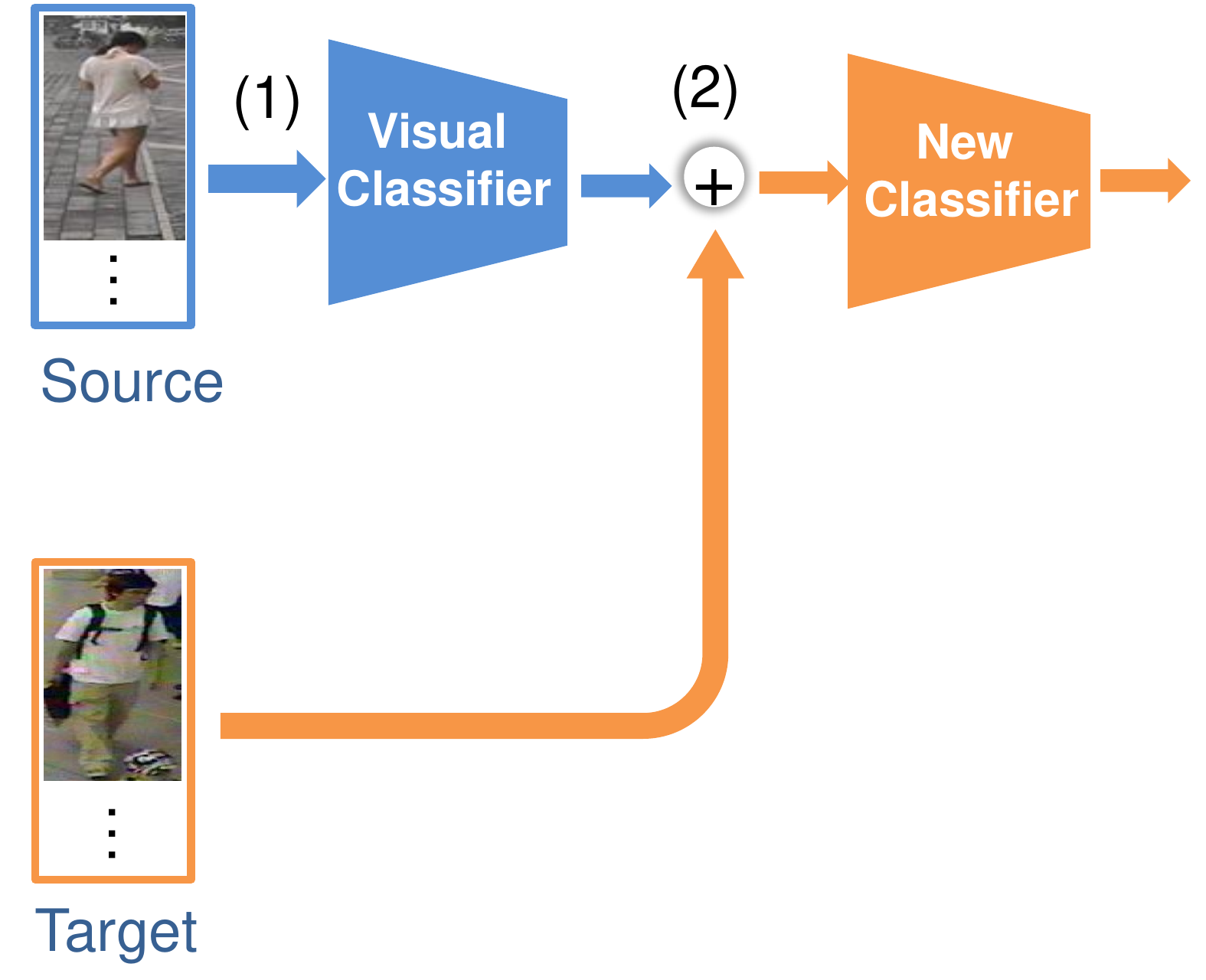}
\end{minipage}
\label{fig:cross-dataset-methods-1}
}
\subfigure[]{
\begin{minipage}[b]{0.3\textwidth}
\includegraphics[width=1.1\textwidth]{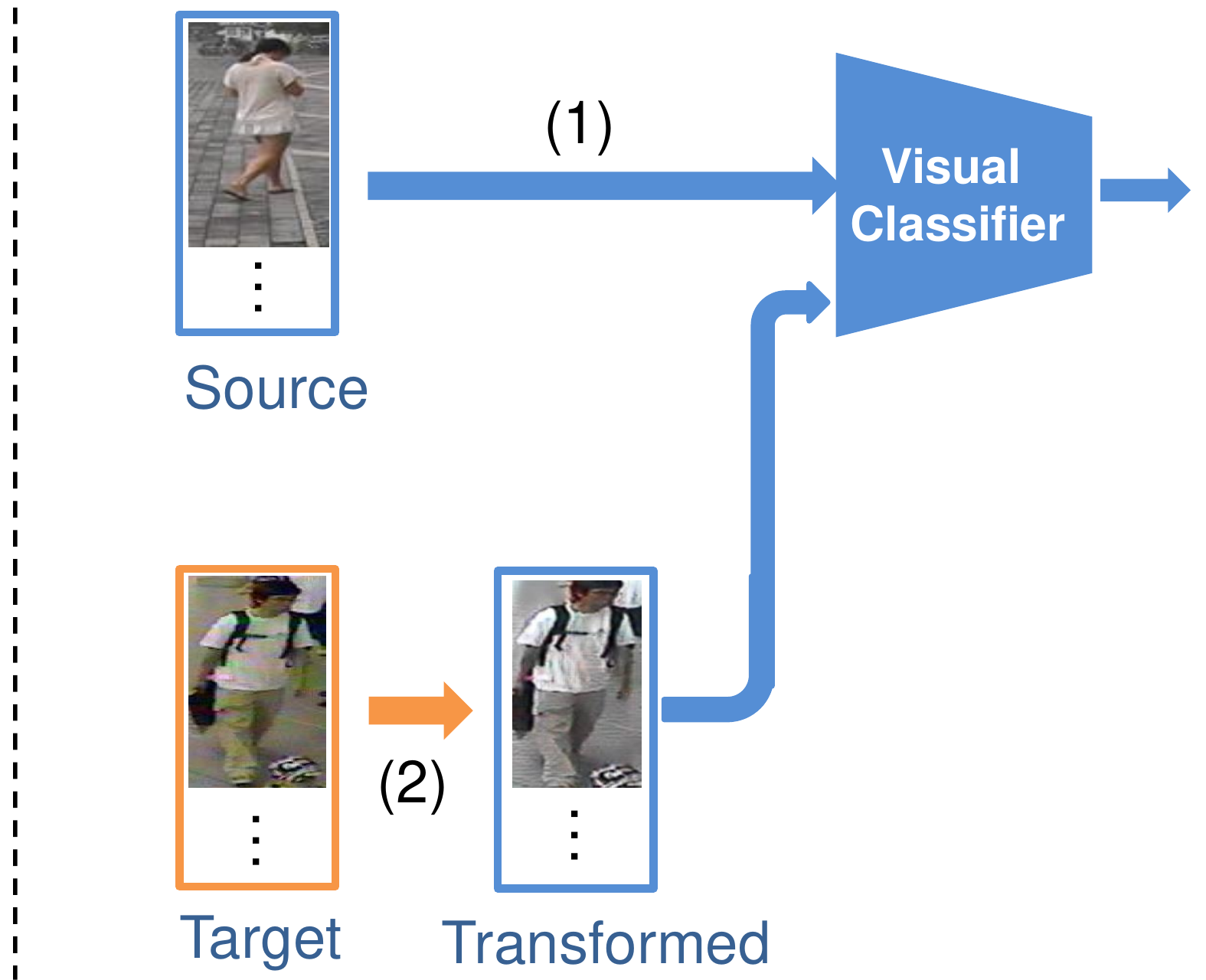}
\end{minipage}
\label{fig:cross-dataset-methods-2}
}
\caption{Two different kind of solutions to perform cross-dataset person Re-ID. (a) Incrementally optimizing the classifier to fit the data in the target domain. Here (1) indicates supervised learning, and (2) indicates incrementally optimization. (b) Transforming the data in the target domain to fit the original classifier. Here (1) indicates supervised learning, and (2) indicates transforming the image into the style of the source domain.}
\label{fig:cross-dataset-methods}
\end{figure}

\section{Model}\label{sec:model}

\subsection{Model overview}

As shown in Fig.1, the images taken in different camera networks usually have significantly different feature distributions. Thus, direct transferring of a visual classifier, which is pre-trained in a small labeled source dataset, to another unlabeled target dataset may cause a poor performance in most cases. As shown in Fig.~\ref{fig:cross-dataset-methods-1}, most of the proposed cross-dataset transfer learning algorithms try to incrementally optimize the classifier based on the unlabeled data from the target dataset. However, without the powerful supervised tuning, the performance of the visual classifier over the target domain usually does not gain significant improvement.

We propose a novel generative model based solution to address the challenge in the opposite way as shown in Fig.~\ref{fig:cross-dataset-methods-2}. Instead of incrementally optimizing the classifier to fit the new data, we transform the data in the target dataset to fit the classifier. In order to make the transformation benefit the improvement of the cross-dataset person Re-ID, the image transformation should satisfy the following constraints:

\begin {itemize}
\item \textbf{Data fitness}. The visual features of the transformed images in the target dataset should fit the feature distribution in the source dataset, and should be more suitable for the visual classifier, which is pre-trained in the source dataset.
\item \textbf{Similarity preservation}.  The similarity of the transformed images, which contain the same person, should be as high as possible.
\end{itemize}

\begin{figure}
\centering
{
\begin{minipage}[b]{0.35\textwidth}
\includegraphics[width=1.2\textwidth]{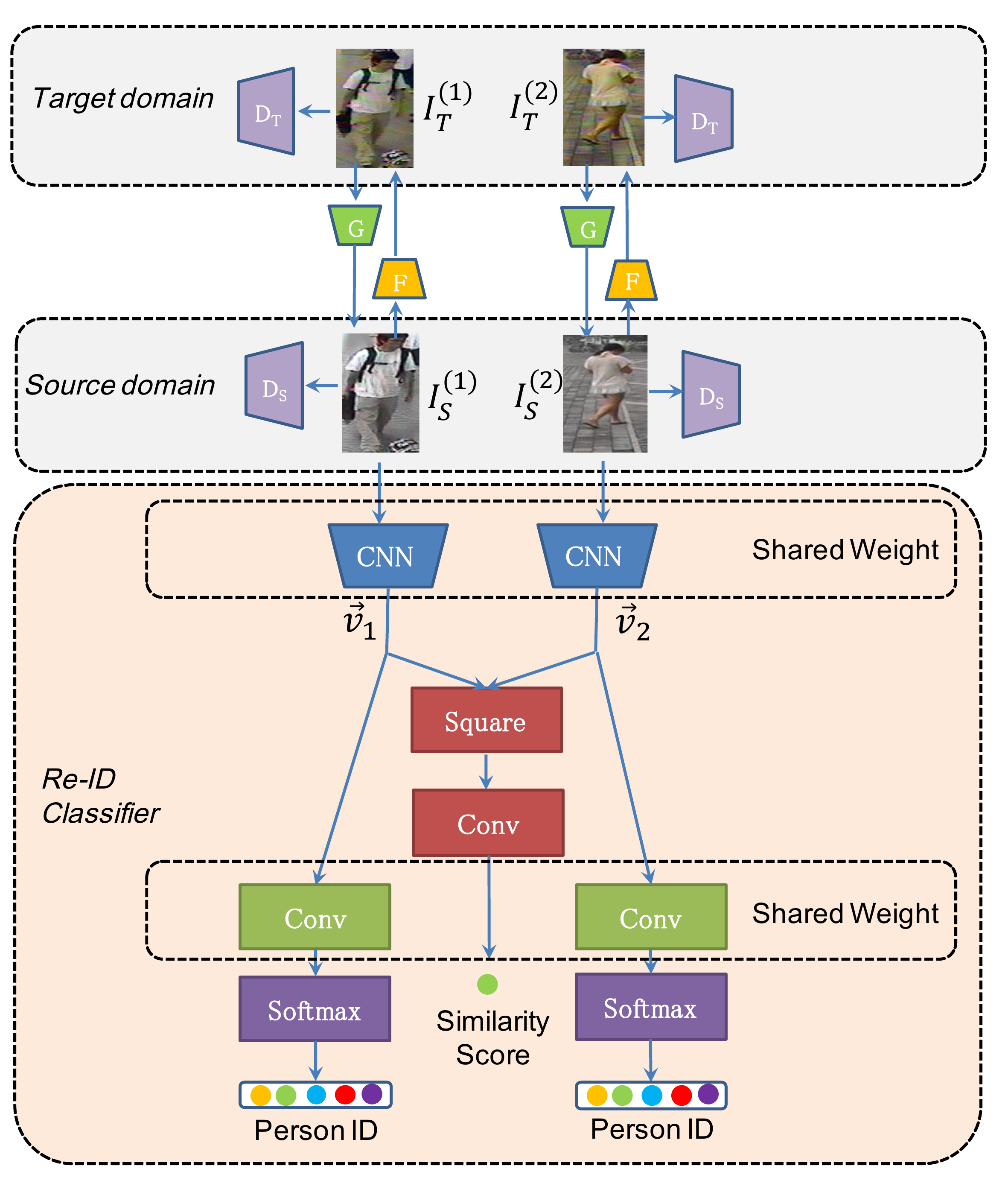}
\end{minipage}
}
\caption{The SimPGAN model for cross-dataset person Re-ID.}
\label{fig:model}
\end{figure}

To address the above constraints, we propose the similarity preserved GAN model, SimPGAN,  to conduct the transformation. As shown in Fig.~\ref{fig:model}, we adopt a siamese deep neural network as the Re-ID classifier, which measures the similarity of two input images and determines the identity of the person in the image. We firstly pre-train this Re-ID classifier in the labeled source dataset, and then we transfer it to the target dataset. Given a pair of images in the tareget dataset, they are simultaneously transformed into the style of the source dataset using the generative model, and input into the Re-ID classifier to measure  their similarity.

To address the `data fitness' constraint,  we integrate the cycle consistency loss\cite{ref_proc29}   in the GAN model, which is proved to be very powerful to improve the quality and steadiness of unpaired images  translation. Meanwhile, to address the `similarity preservation' constraint, we use the similarity loss from the Re-ID classifier as the high-level semantic signal to fine-tune the generative model to preserve the similarity of the transformed images of the same person.  We will introduce the detail of each component in the following sections.

\subsection{Siamese CNN based Re-ID Classifier}
As shown in Fig.~\ref{fig:model}, we select the recently proposed siamese convolutional  neural network \cite{ref_proc28}  as the Re-ID classifier, which makes better use of the label information and has a good performance in the large-scale datasets such as Market1501\cite{ref_proc27} . The network adopts a siamese scheme including two ImageNet pre-trained CNN modules, which share the same weight parameters and extract visual features from the input images $I_S^{(1)}$ and $I_S^{(2)}$. The CNN module is achieved from the ResNet-50 network \cite{ref_proc6}  by removing its final fully-connected (FC) layer. The outputs of the two CNN modules are flattened into two one-dimensional vectors: $\vec{v_1}$ and $\vec{v_2}$, which act as the embedding visual feature vectors of the input images.

To measure the matching degree of the input images, their feature vectors $\vec{v_1}$ and $\vec{v_2}$ are fed into the following square layer to conduct subtracting and squaring element-wisely: $\vec{v_s} = (\vec{v_1} - \vec{v_2})^2$. Finally, a convolutional layer is used to transform $\vec{v_s}$ into the similarity score as:
\begin{eqnarray}
\hat{q} = sigmoid(\theta_s \circ \vec{v_s})
\end{eqnarray}
Here $\theta_s$ denotes the parameters in the convolutional layer, $\circ$ denotes the convolutional operation, and $sigmoid$ indicates the $sigmoid$ activation function. By comparing the predicted similarity score with the ground-truth matching result of $I_S^{(1)}$ and $I_S^{(2)}$, we can achieve the \textbf{variation loss} as a cross entropy form:

\begin{footnotesize}
\begin{eqnarray}
\emph{L}_v(I_S^{(1)}, I_S^{(2)}) = -q \cdot log(\hat{q}) - (1 - q) \cdot log(1-\hat{q})
\end{eqnarray}
\end{footnotesize}
Here $q = 1$ when $I_S^{(1)}$ and $I_S^{(2)}$ contain the same person. Otherwise, $q = 0$.

Besides predicting the similarity score, the model also predicts the identity of each image in the following steps.  Each visual feature vector ($\vec{v_x} (x=0,1)$ ) is  fed into one convolutional layer to be mapped into an one-dimensional vector with the size $K$, where $K$ is equal to the total number of the pedestrians in the dataset. Then the following softmax unit is applied to normalize the output as follows:
\begin{eqnarray}
\hat{P}^{(x)} = softmax ( \theta_x \circ \vec{v_x}) (x=1,2)
\end{eqnarray}
Here $\theta_x$ is the parameter in the convolutional layer and $\circ$ denotes the convolutional operation. The output  $\hat{P}^{(x)}$ is used to predict the identity of the person contained in the input image $I_S^{(x)}(x=1,2)$. By comparing $\hat{P}^{(x)}$ with the ground-truth identify label, we can achieve the \textbf{identification loss} as the cross-entropy form:

\begin{footnotesize}
\begin{eqnarray}
&&L_{id}(I_S^{(1)}, I_S^{(2)}) \nonumber \\
 &=&\sum_{k=1}^{K}(-log \hat{P}^{(i)}_k \cdot P^{(i)}_k) + \sum_{k=1}^{K}(-log \hat{P}^{(j)}_k \cdot P^{(j)}_k)
\end{eqnarray}
\end{footnotesize}
Here $P^{(x)} (x=1,2)$ is the identity vector of the input image $I_S^{(x)}$. $P^{(x)}_k = 0$ for all $k$  except $P^{(x)}_t = 1$, where $t$ is ID of the person in the image $I_S^{(x)}$.

While training the classifier in the labeled source dataset, the parameters are optimized by using the final loss function:
\begin{footnotesize}
\begin{eqnarray}
L_{all} = L_v + L_{id}
\end{eqnarray}
\end{footnotesize}
According to \cite{ref_proc24} , this kind of  composite loss makes the classifier more efficient to extract the view-invariant visual features for Re-ID than the single loss function.

\subsection{Similarity Preserved GAN}
While transferring the Re-ID classifier to the target domain, we propose a siamese GAN archtecture to transform each pair of the images in the target domain to the style of the source domain and feed them into the Re-ID classifier as shown in Fig.~\ref{fig:model}. Similar with the implementation of CycleGAN\cite{ref_proc29} , the model includes two generators performing the cross-dataset image mapping: $G:I_S \rightarrow I_T$ and $F:I_T \rightarrow I_S$, where $I_S$ indicates the source dataset and $I_T$ indicates the target dataset. In addition, the model integrates the adversarial discriminator classifier $D_S$ to discriminate  the real images $I_S$ and generated images $F(I_T)$,  and $D_T$ to discriminate  the real images $I_T$ and generated images $G(I_S)$.

The optimization of $G$,$F$,$D_S$ and $D_T$ is based on the combination of the  cycle consistency adversarial loss\cite{ref_proc29}   and our proposed similarity consistency loss, which will be introduced in the following sections. As shown in Fig.~\ref{fig:model}, in each iteration of optimization, the model takes four images as input to calculate the loss: one pair of labeled images ($I_S^{(1)}$,$I_S^{(2)}$) from the source dataset, and another pair of unlabeled images ($I_T^{(1)}$,$I_T^{(2)}$) from the target dataset.

\subsubsection{Cycle Consistency Adversarial Loss}

We apply the recently proposed least square adversarial loss \cite{ref_proc18} to optimize each pair of generator and discriminator. Specifically , the loss for the generator $G:I_S \rightarrow I_T$ is defined as:

\begin{footnotesize}
\begin{eqnarray}
L_G = \sum_{x=1,2}(D_T(G(I_S^{(x)}))-1)^2
\end{eqnarray}
\end{footnotesize}

Minimizing the loss means to transform the image $I_S^{(x)}(x = 1,2)$ in the source domain to make the discriminator $D_T$ believe it is an image from the target domain. Similarly, we can define the loss for the discriminator $D_T$ as:

\begin{footnotesize}
\begin{eqnarray}
L_{D_T} = \sum_{x=1,2}[(D_T(G(I_S^{(x)})))^2 + (D_T(I_S^{(x)})-1)^2]
\end{eqnarray}
\end{footnotesize}
Here the discriminator $D_T$  tries to distinguish the real image $I_S^{(x)}$ and the transformed image $G(I_S^{(x)})$. Obviously, $L_G$ and $L_{D_T}$ are a pair of adversarial loss. In a similar way, we can define the loss for the generator $F:I_T \rightarrow I_S$ as:
\begin{footnotesize}
\begin{eqnarray}
L_F = \sum_{x=1,2}(D_S(F(I_T^{(x)}))-1)^2
\end{eqnarray}
\end{footnotesize}
and the loss of the discriminator $D_S$ as:
\begin{footnotesize}
\begin{eqnarray}
L_{D_S} = \sum_{x=1,2}[(D_S(F(I_T^{(x)})))^2 + (D_S(I_T^{(x)})-1)^2]
\end{eqnarray}
\end{footnotesize}

Similar to the CycleGAN\cite{ref_proc29} , we also integrate the cycle consistency loss into the model to avoid sharp change the images:
\begin{footnotesize}
\begin{eqnarray}
L_{cycle} \nonumber = \sum_{x=1,2} [\|G(F(I_T^{(x)})) - I_T^{(x)} \|_1 + \|F(G(I_S^{(x)})) - I_S^{(x)} \|_1] \nonumber
\end{eqnarray}
\end{footnotesize}

\subsubsection{Similarity Consistency Loss}

We propose the similarity consistency loss to address the `similarity preservation' constraint as:
\begin{footnotesize}
\begin{eqnarray}
L_{sim} = L_v (F(G(I_S^{(1)})), F(G(I_S^{(2)})))
\label{eq:sim-loss}
\end{eqnarray}
\end{footnotesize}
Here $L_v$ denotes the variation loss of the two images in the source domain as defined in Eq.~(2), which indicates the loss of the similarity measurement of the images based on their ground-truth labels.  $F(G(I_S^{(x)}))(x=0,1)$ indicates transforming the image $I_S^{(x)}$ in the source domain  to the style of the target domain, and then further transforming back into the style of the source domain. Minimizing this similarity consistency loss (Eq.~(\ref{eq:sim-loss})) means making the similarity of the transformed images containing the same person as high as possible, and thus preserving the precision of the similarity measurement during the transformation.

\subsection{Optimization procedure}
We adopt the Stochastic Gradient Descent method to optimize the generators ($G$,$F$) and the discriminators ($D_S$, $D_T$). Specifically, we integrate all  of the generators related loss as:
\begin{footnotesize}
\begin{eqnarray}
L_{gen} = L_G + L_F + \lambda_1 L_{cycle} + \lambda_2 L_{sim}
\label{eq:total-loss}
\end{eqnarray}
\end{footnotesize}
where  $\lambda_1$ and $ \lambda_2$ control the relative importance of each loss. We also combine all of the discriminators related loss as:
\begin{footnotesize}
\begin{eqnarray}
L_{dis} = L_{D_S} + L_{D_T}
\end{eqnarray}
\end{footnotesize}
In each training iteration, we use the loss $L_{gen}$ to optimize the parameters of $G$ and $F$, and use $L_{dis}$ to optimize the parameters of $D_S$ and $D_T$ by gradient descent.

\section{Experiment}\label{sec:experiment}

\subsection{Dataset and Setting}

Three widely used benchmark datasets are chosen in our experiments, including GRID \cite{ref_proc2} , Market1501 \cite{ref_proc27} , CUHK01 \cite{ref_proc23} . As shown in the Table.~\ref{tab:comparison}, we select one of the above datasets as the source dataset and another one as the target dataset to test the performance of  cross-dataset person Re-ID.  In each source dataset, all the labeled images are used for the training of the visual classifier. On the other hand, the configurations of the target dataset `GRID' follow the instructions of \cite{ref_proc2}  to divide the training and testing set, and  a 10-fold cross validation is conducted.

We compare the performance of the following cross-dataset person Re-ID algorithms in the experiments:
\begin{itemize}
\item \textbf{UMDL} \cite{ref_proc18} : It is the state-of-the-art unsupervised cross-dataset person Re-ID algorithm. UMDL is based on the cross-dataset dictionary learning method and outperforms other unsupervised learning methods as reported in \cite{ref_proc18} .
\item \textbf{Direct Transfer}: It is the baseline model, which directly transfers the siamese neural network based Re-ID classifier (as shown in Fig.~\ref{fig:model}), which is pre-trained in the source dataset, to the target dataset.
\item \textbf{GAN}: It is the original version of GAN, which only uses the adversarial loss ($L_G$,$L_F$,$L_{D_S}$ and $L_{D_T}$)  to train the model.
%\item \textbf{SimPGAN(G+C)}: It is the simplified version of SimPGAN, which uses the adversarial loss with the cycle consistency loss $L_{cycle}$   to train the model.
\item \textbf{SimPGAN}: It is the SimPGAN model, which integrates the adversarial loss, the cycle consistency loss and the similarity consistency loss $L_{sim}$. $\lambda_1$ and $\lambda_2$ are two tunable parameters in Eq.~(\ref{eq:total-loss}). By default, we set $\lambda_1=10$ and $\lambda_2=1$.

\end{itemize}

\begin{table}[!htbp]   \scriptsize
\centering
\caption{Cross-dataset re-id precision.}\label{tab:umdl}
\begin{tabular}{l|l|l|ccc}
\toprule
 \multirow{2}*{Method}&\multirow{2}*{Source}&\multirow{2}*{Target}& \multicolumn{3}{c}{Performance}\\
\cline{4-6}
 &&& rank-1& rank-5& rank-10\\
\hline
\hline
\multirow{3}*{UMDL}
&Market1501 & GRID & 3.77 & 7.76 &  9.71\\
&CUHK01 & GRID & 3.58 & 7.56 &  9.50\\
%&GRID& Market1501 & 3.97 & 8.14 &  10.73\\
\hline
\multirow{3}*{Direct Transfer}
&Market1501 & GRID & 9.60 & 21.20 &  28.40\\
&CUHK01 & GRID & 3.60 & 7.20 &  9.20\\
%&GRID& Market1501 & 7.19 & 16.86 &  22.77\\
\hline
\multirow{3}*{GAN}
&Market1501& GRID&\multicolumn{1}{|c}{ 4.40} & 10.80 & 15.20\\
&CUHK01& GRID&\multicolumn{1}{|c}{2.40} & 6.40 & 11.60\\
%&GRID& Market1501&\multicolumn{1}{|c}{2.26} & 6.98 & 10.21\\
\hline
\multirow{3}*{SimPGAN}
&Market1501& GRID&\multicolumn{1}{|c}{ \textbf{18.00}} & \textbf{34.40} & \textbf{43.20}\\
&CUHK01& GRID&\multicolumn{1}{|c}{\textbf{13.20}} & \textbf{24.40} & \textbf{32.40}\\
%&GRID& Market1501&\multicolumn{1}{|c}{8.93} & 20.90 & 27.70\\
\bottomrule
\end{tabular}
\label{tab:comparison}
\end{table}

\begin{figure}
\centering
{
\begin{minipage}[b]{0.45\textwidth}
\includegraphics[width=1\textwidth]{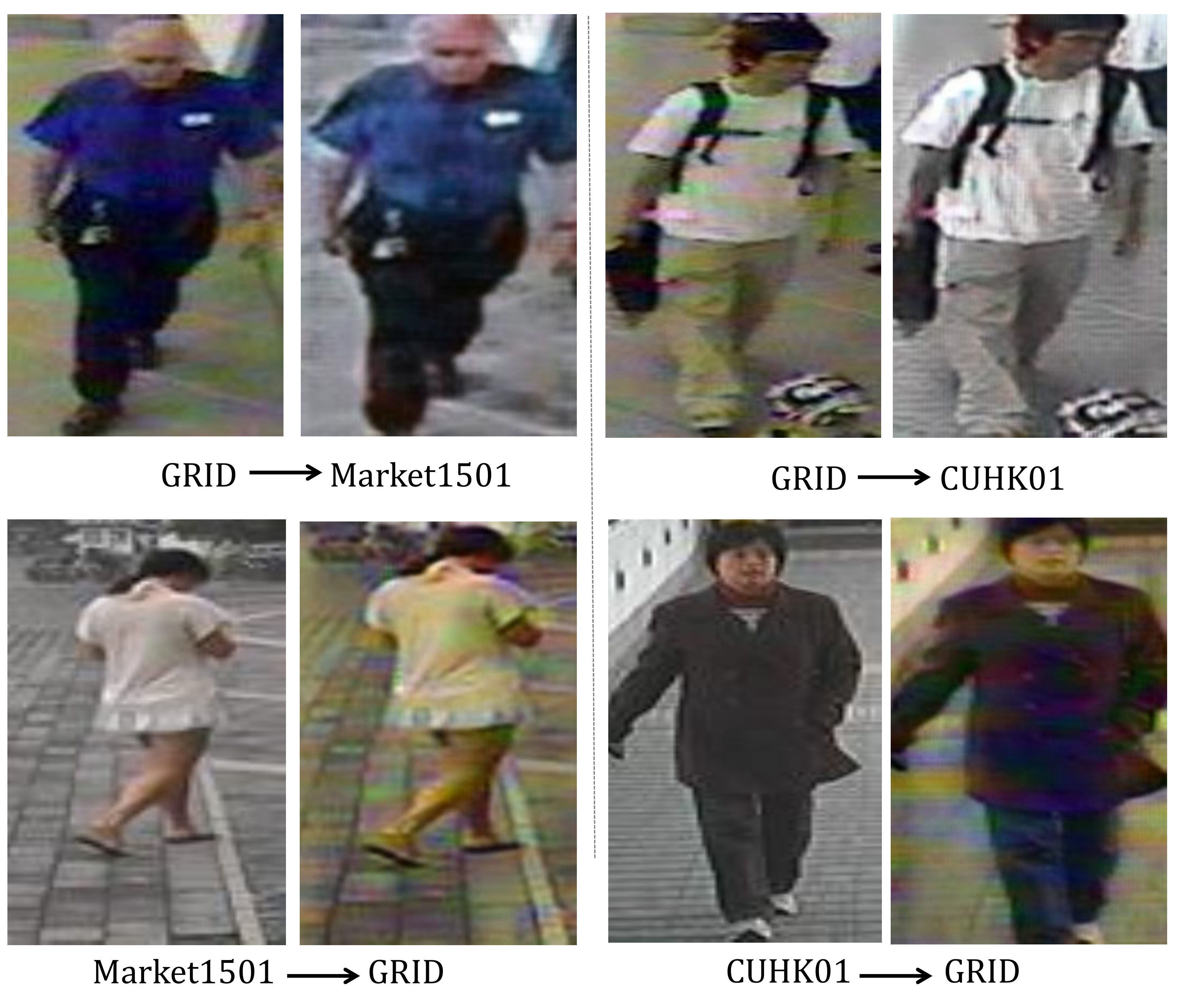}
\end{minipage}
}
\caption{Transform the images in the source dataset to the style of the target dataset.}
\label{fig:transform-image}
\end{figure}
\subsection {Re-ID Results}

We compare the performance of different cross-dataset transfer learning algorithms in Table.~\ref{tab:comparison}.  The performance of `Direct Transfer' baseline model is quite poor, which is due to the different styles of the source and target datasets as shown in Fig.1. The `SimPGAN' model gains a lot of improvement compared with the    `Direct Transfer' model, and also outperforms UMDL\cite{ref_proc18}  with a large margin. This proves that the image transformation greatly improves the fitness of the target data to the classifier transferred from the source domain. To verify the effectiveness  of each component in SimPGAN, we also test the original version of GAN. Without the constraint of the cycle consistency loss and similarity consistency loss, `GAN' achieves a much worse performance. This confirms the importance of the cycle consistency loss and the similarity consistency loss to preserve the identity information of the transformed images.

\begin{figure}[!htbp]
\centering
\subfigure[]{
\begin{minipage}[b]{0.3\textwidth}
\includegraphics[width=1\textwidth]{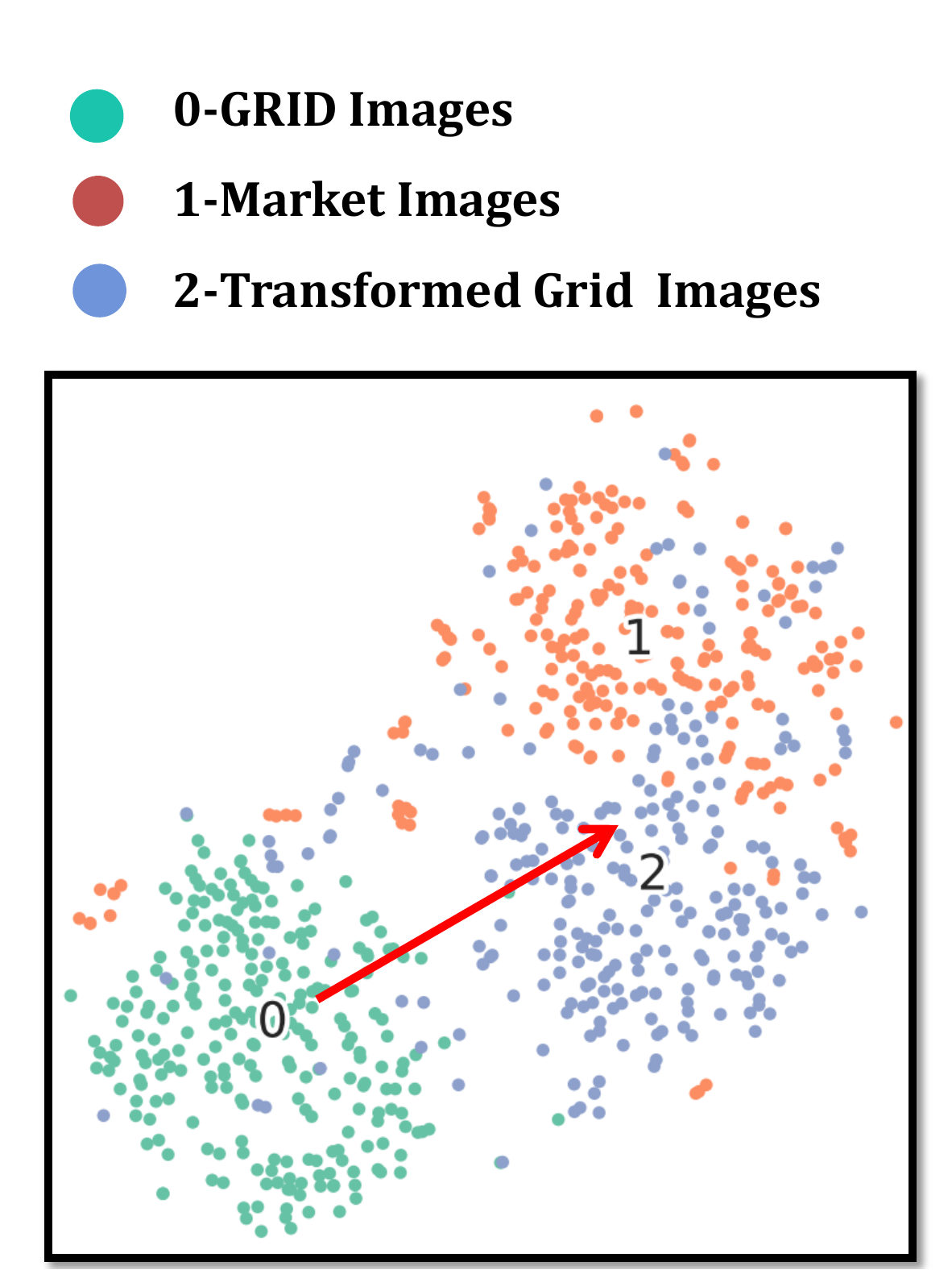}
\end{minipage}
\label{fig:feature_transform_all_1}
}
\subfigure[]{
\begin{minipage}[b]{0.3\textwidth}
\includegraphics[width=1\textwidth]{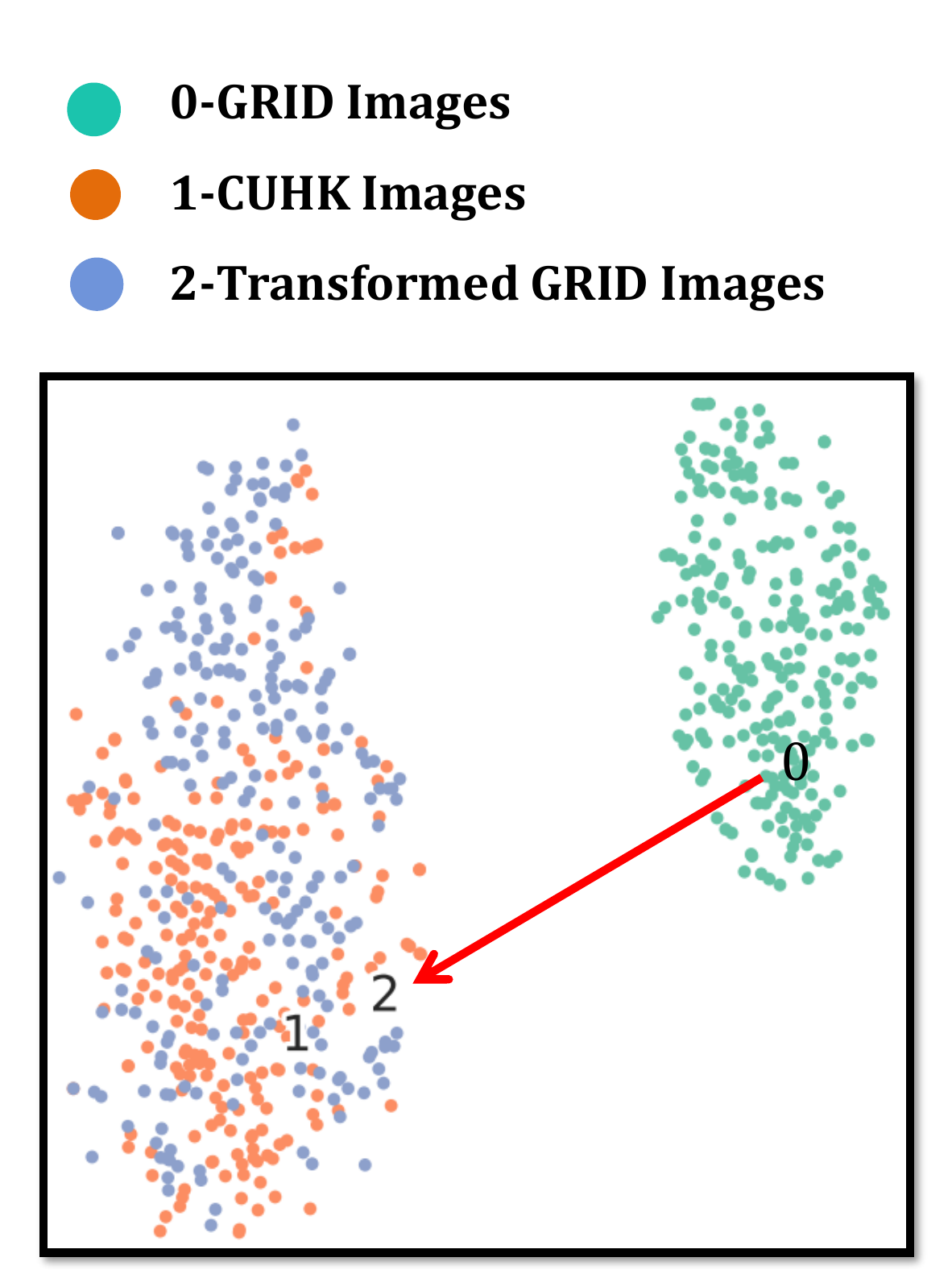}
\end{minipage}
\label{fig:feature_transform_all_2}
}
\caption{Feature distribution of the images. (a)Transforming GRID to Market1501. (b) Transforming GRID to CUHK01.}
\label{fig:feature_transform_all}
\end{figure}
To further observe how SimPGAN transforms images, we compare the transformed images with the original ones in  Fig.~\ref{fig:transform-image}. It shows clearly that the images in the target domain are transformed into the style of the source domain. Furthermore, Fig.~\ref{fig:feature_transform_all} shows how the  distribution of the visual features of the images changes after transformation. The feature vector of each image is extracted by the siamese convolutional neural network, which is trained on the source dataset, and is corresponding to  $\vec{v_i}(i=1,2)$ in Fig.~\ref{fig:model}. Each feature vector is transformed into a 2-dimensional embedding vector by the t-SNE \cite{ref_proc7}  algorithm and plotted in Fig.~\ref{fig:feature_transform_all}. It shows that the feature vectors of the transformed images in the target dataset are much closer to those in the source dataset after the transformation.

\section{Conclusions} \label{sec:conclusions}

In this paper, we have presented SimPGAN as an efficient unsupervised cross-dataset person Re-ID algorithm based on image transformation. In particular, SimPGAN transforms the unlabeled images in the target dataset into the style of the source dataset to fit the siamese convolutional classifier, which is pre-trained on the labeled source dataset. By integrating with the similarity consistency loss and cycle consistency loss, SimPGAN preserves the important identity information of the images during the transformation. Experiments show that  SimPGAN can greatly improve the fitness of unlabeled target dataset to the Re-ID classifier transferred from the source dataset, and outperforms the state-of-the-art unsupervised cross-dataset transferring algorithm by a big margin.

Furthermore, SimPGAN is easy to use as an image enhancement tool to improve the performance of other cross-dataset person Re-ID algorithms. In our future work, we will integrate the image transformation of SimPGAN with other cross-dataset person Re-ID algorithms to enhance their performance.


\end{document}